\documentclass[10pt,twocolumn,letterpaper]{article}

\usepackage{iccv}              

\definecolor{iccvblue}{rgb}{0.21,0.49,0.74}
\usepackage[pagebackref,breaklinks,colorlinks,allcolors=iccvblue]{hyperref}
\usepackage{makecell}
\def\paperID{4373} 
\def\confName{ICCV}
\def\confYear{2025}

\title{Towards Generalization of Tactile Image Generation: Reference-Free Evaluation in a Leakage-Free Setting}

\newcommand*{\affaddr}[1]{#1} 
\newcommand*{\affmark}[1][*]{\textsuperscript{#1}}

\author{%
Cagri Gungor\affmark[1], Derek Eppinger\affmark[2] and Adriana Kovashka\affmark[1,2]\\
\affaddr{\affmark[1]Intelligent Systems Program}, \affaddr{\affmark[2]Department of Computer Science}\\
\affaddr{University of Pittsburgh}\\
}

\begin{document}
\maketitle

\begin{abstract}
Tactile sensing, which relies on direct physical contact, is critical for human perception and underpins applications in computer vision, robotics, and multimodal learning. Because tactile data is often scarce and costly to acquire, generating synthetic tactile images provides a scalable solution to augment real-world measurements. However, ensuring robust generalization in synthesizing tactile images—capturing subtle, material-specific contact features—remains challenging. We demonstrate that overlapping training and test samples in commonly used datasets inflate performance metrics, obscuring the true generalizability of tactile models. To address this, we propose a leakage-free evaluation protocol coupled with novel, reference-free metrics—TMMD, I-TMMD, CI-TMMD, and D-TMMD—tailored for tactile generation. Moreover, we propose a vision-to-touch generation method that leverages text as an intermediate modality by incorporating concise, material-specific descriptions during training to better capture essential tactile features. Experiments on two popular visuo-tactile datasets, Touch and Go and HCT, show that our approach achieves superior performance and enhanced generalization in a leakage-free setting.
\end{abstract}

\section{Introduction}
\label{sec:intro}

The tactile modality, which relies on direct physical contact to sense attributes such as force, pressure, and temperature, 
is a unique sensory channel that is fundamental to our understanding of the world \cite{hutmacher2019there, linden2016touch}. In computer vision and robotics, tactile cues effectively complement visual data in multimodal learning, enabling systems to discern subtle material properties that pure vision might overlook \cite{cao2024multimodal, yuan2018active, cao2020spatio, kerr2022self}. This integration not only provides critical feedback for robotic manipulation through nuanced contact dynamics \cite{yuan2024robot, guzey2023dexterity, lloyd2021goal}, but also enriches scene understanding by augmenting 3D representations with tactile insights \cite{dou2024tactile}. However, obtaining robust and generalizable models remains a significant challenge across various tactile sensing tasks due to the expensive physical data collection process and non-standardized sensor outputs \cite{yang2024binding}. 

\begin{figure}[t]
    \centering
    \includegraphics[width=1\linewidth]{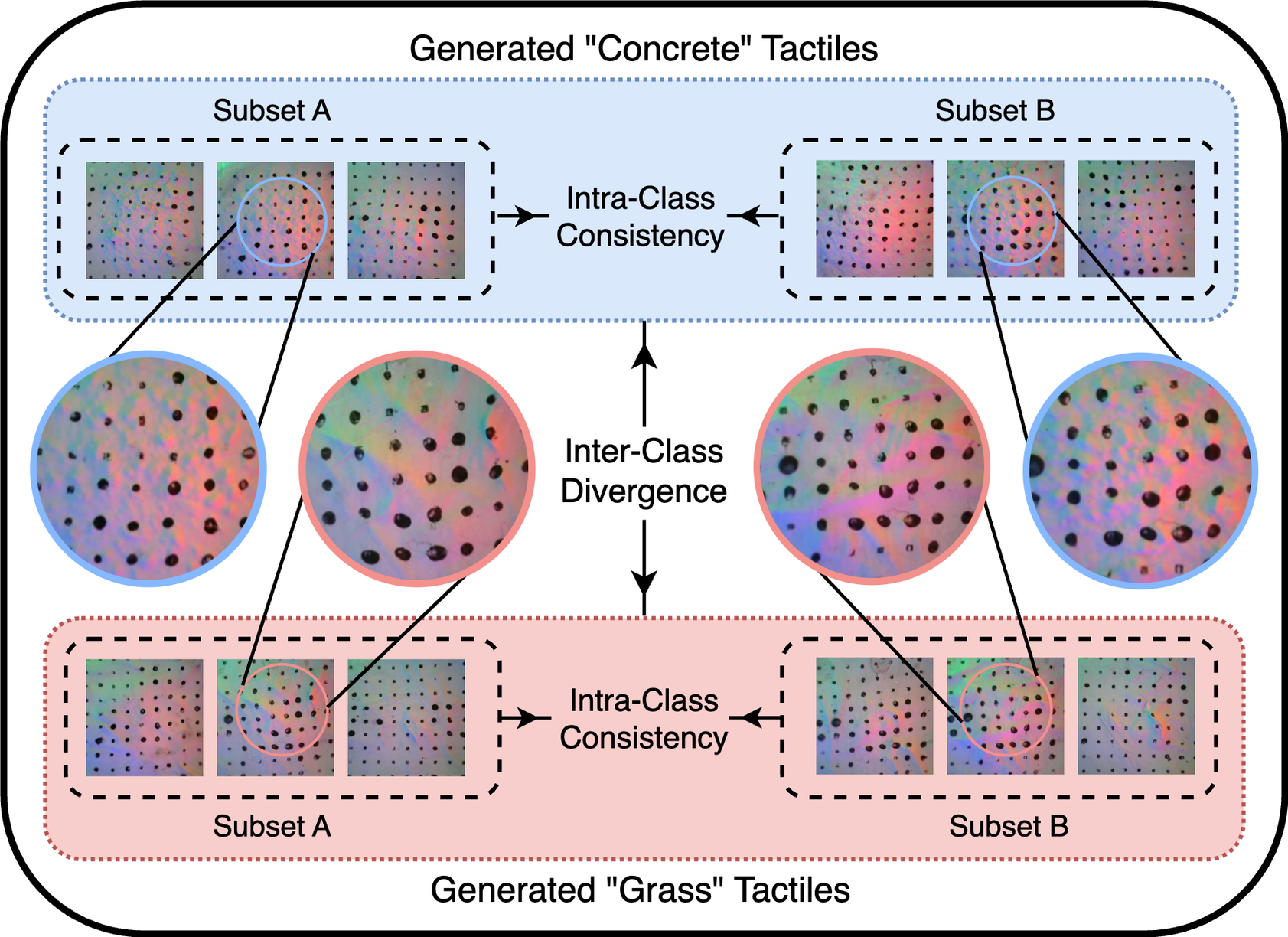}    
    \caption{An illustration of how our reference-free metrics evaluate generated tactile images across different classes. Within each class (e.g., “Concrete” or “Grass”), samples are split into two subsets (A and B) to measure \emph{intra-class consistency}, while \emph{inter-class divergence} quantifies how distinct the generated samples are between classes.} 
    \label{fig:concept}
\end{figure}

Tactile generation stands out as a promising avenue for synthesizing tactile images from visual and multimodal inputs, offering a potential solution to mitigate the scarcity of tactile data. Recent research \cite{yang2023generating,dou2024tactile,tu2024texttoucher,rodriguez2024touch2touch,gao2023controllable} in this area has demonstrated encouraging progress; however, the evaluation of these models remains problematic for two following reasons: data leakage and lack of specialized metrics.

Data leakage poses a critical obstacle to advancing tactile sensing research, undermining the goal of robust generalization by compromising many prevalent tactile datasets \cite{yang2022touch, fu2024touch, kerr2022self}. These datasets, often derived from continuous video recordings of object interactions, inadvertently include overlapping or near-duplicate samples across training and test splits.
This leakage enables models to simply memorize redundant tactile patterns. 
It also results in artificially inflated performance metrics across various tactile-related tasks that do not reflect real-world robustness. Consequently, current evaluations mask the true challenges of different tasks such as generating high-fidelity tactile images that capture the subtle, material-specific features. As a first contribution, we thoroughly analyze the leakage phenomenon—detailing its causes and impact—and introduce a leakage-free evaluation protocol that rigorously separates training and test subsets, thereby enabling more reliable and generalizable results, meaning consistent performance across diverse, unseen scenarios to address data variability.

In evaluating tactile image generation, many works have adopted conventional metrics—such as SSIM, PSNR, LPIPS, and FID—that were originally designed for natural images. These metrics often fail to accurately capture tactile-specific nuances on generated tactile images. They exhibit excessive sensitivity to irrelevant background artifacts due to passive sensor state,  in the absence of contact to a material. Meanwhile they neglect fine, material-specific features that are essential for tactile perception. This misrepresentation skews performance evaluations and obscures the true generalizability of the models. As a second contribution, we perform an in-depth analysis of the limitations of these metrics and we introduce a set of novel, reference-free evaluation metrics—namely, TMMD, I-TMMD, CI-TMMD, and D-TMMD—which are specifically designed to capture material-specific tactile features and eliminate the need for reference images. By leveraging a dedicated tactile encoder, these metrics quantify both intra-class consistency (ensuring internal stability) and inter-class divergence (ensuring diversity across materials) as illustrated in Fig.~\ref{fig:concept}, thereby offering an evaluation to accurately assess tactile image fidelity and diversity of generated images. This comprehensive evaluation ultimately fosters the development of more reliable and generalizable tactile generation systems.

Finally, we propose a method that leverages text as an intermediate modality in the vision-to-touch generation task. This is achieved by incorporating concise, material-specific textual descriptions during training. By leveraging text to emphasize essential tactile features and filter out irrelevant visual details, our method accurately captures intrinsic contact patterns and material properties, thereby enhancing the quality of the generated tactile images. We use our robust evaluation framework, which operates under an enforced leakage-free setting and utilizes novel metrics tailored specifically for tactile generation, to verify the effectiveness of our method, ensuring that the results are reliable and generalizable.
\section{Related Work}
\label{sec:related_work}

\textbf{Tactile sensing and visual-tactile datasets.}
In recent years, various tactile sensors have been developed and extensively utilized in robotics \cite{zhao2024tactile,calandra2017feeling,yuan2024robot,calandra2018more,feng2025learning}. Traditional sensors \cite{cutkosky2016force,kappassov2015tactile} captured low-dimensional signals such as pressure, force, and vibration, providing limited surface details. To further enhance robotic interaction and manipulation, vision-based tactile sensors, such as GelSight \cite{yuan2017gelsight}, DIGIT \cite{lambeta2020digit}, and Taxim \cite{si2022taxim}, overcome this by using elastomeric materials with embedded cameras and lighting for high-resolution tactile imaging. Many visual-tactile datasets, such as Touch and Go \cite{yang2022touch}, HCT \cite{fu2024touch}, SSVTP \cite{kerr2022self}, VisGel \cite{li2019connecting} and TaRF \cite{dou2024tactile} have been developed using vision-based tactile sensors to support various computer vision tasks, including self-supervised representation learning \cite{yang2024binding,fu2024touch}, material classification \cite{yang2022touch, yang2024binding}, tactile-driven image stylization \cite{yang2024binding,fu2024touch}, and cross-modal image generation \cite{dou2024tactile, yang2023generating, tu2024texttoucher}. However, our analysis reveals that some of these datasets include highly similar samples in both training and test sets, leading to overfitting and inflated performance metrics that poorly reflect real-world generalization.

\noindent
\textbf{Cross-modal generation.}
Humans naturally connect information across different senses, a remarkable ability that has inspired extensive research in cross-modal generation. Early works laid the foundation by developing generative frameworks that could convert images from one representation to another \cite{isola2017image,liu2017unsupervised,zhu2017unpaired,wang2018high}. More recently, diffusion‐based methods have become popular in cross‐modal image translation because they tend to produce high‐quality, photorealistic images with stable training dynamics \cite{dhariwal2021diffusion,rombach2022high}. These advances have been extended to a variety of conditioning modalities, enabling impressive results in tasks such as text‐to‐image generation \cite{khachatryan2023text2video,xu2023versatile, xue2024raphael,aghazadeh2024cap}, audio‐to‐image synthesis  \cite{Sung-Bin_2023_CVPR,biner2024sonicdiffusion}, and video‐to‐audio translation  \cite{luo2024diff}. In the field of tactile generation, Touch2Touch \cite{rodriguez2024touch2touch} facilitates translation between various tactile sensor outputs. While TextToucher \cite{tu2024texttoucher} focuses on text-to-touch generation, GVST \cite{yang2023generating} and UniTouch \cite{yang2024binding} employ cross-modal synthesis between visual and tactile images, using similar architectures. Yet these approaches are still constrained by dataset issues (data leakage) that compromise evaluation reliability. In contrast, our work adopts a leakage-free setting and integrates text as an intermediate modality during training for vision-to-touch generation, thereby enhancing the model’s ability to capture material-specific tactile cues and ensuring robust generalization.

\noindent
\textbf{Evaluation metrics for generated images.}
Recent works in image generation evaluation employ various metrics, each capturing different aspects of image quality. Traditional measures like SSIM \cite{wang2004image} and PSNR are pixel-based, making them overly sensitive to minor spatial misalignments. LPIPS \cite{zhang2018unreasonable} improves on this by comparing learned feature representations but lacks distributional awareness. In contrast, FID \cite{heusel2017gans} and KID \cite{binkowski2018demystifying} assess distributional similarity, capturing higher-level semantics. The recent CMMD \cite{jayasumana2024rethinking} leverages CLIP embeddings with MMD, offering a more robust and sample-efficient alternative to FID. However, all these metrics rely on reference images, which can be impractical  for domains where obtaining reference is costly or infeasible. To address this, we propose novel metrics tailored for tactile image generation, incorporating a touch-specific encoder and both reference-free and class-conditioned evaluation strategies.


\section{Method}
\label{sec:method}

    
 

\begin{figure}[t]
    \centering
    \includegraphics[width=1\linewidth]{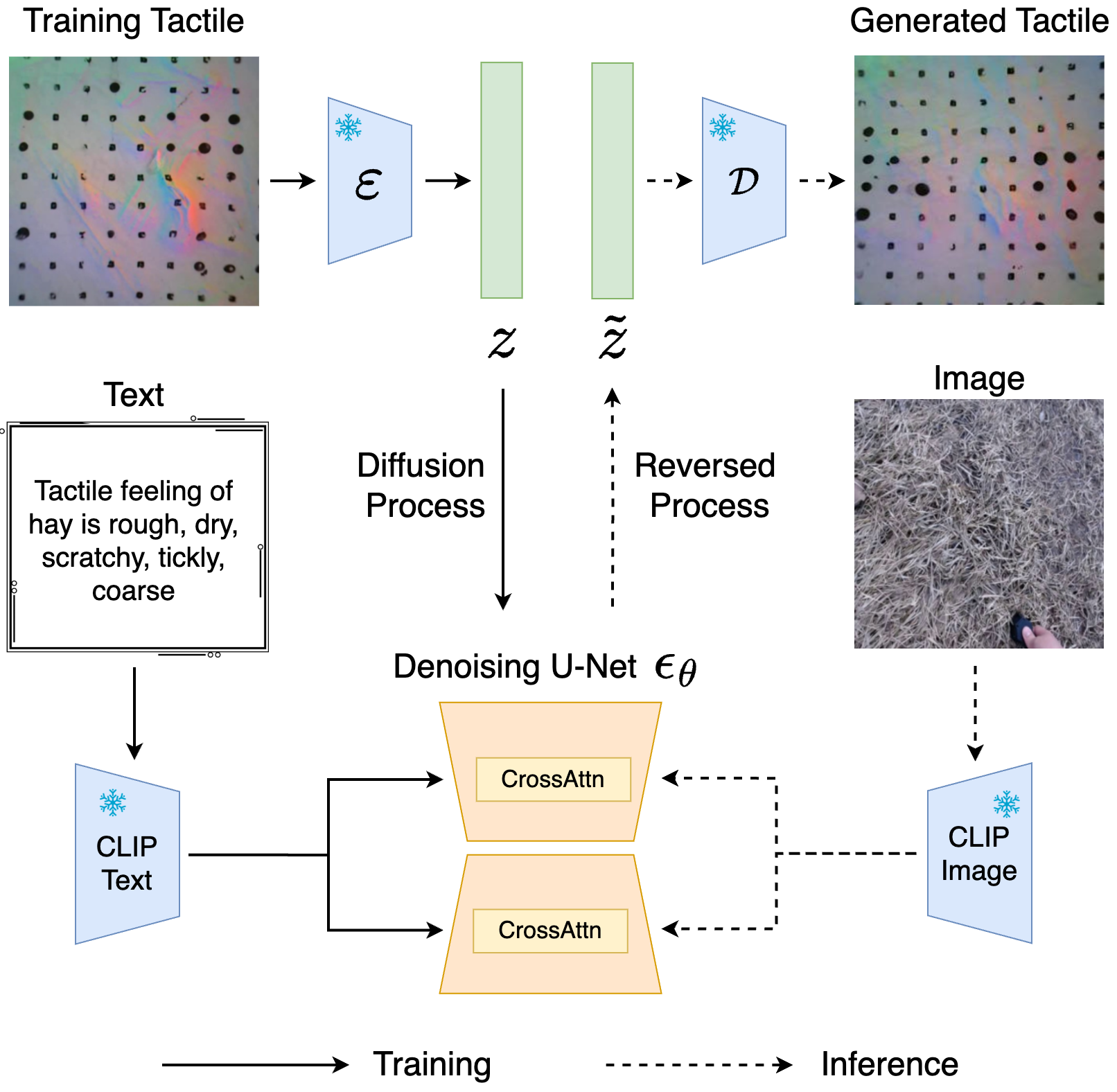}
    \caption{Overview of our latent diffusion-based tactile image generation pipeline: visual inputs are enriched with text cues during training to guide the generation of tactile images via latent diffusion. At inference, only the visual image is required.}
    \label{fig:diff}
    
\end{figure}

Due to the labor-intensive process of collecting tactile data, we propose an approach that generates tactile images from visual inputs to alleviate data scarcity. We leverage textual descriptions of tactile properties as an intermediate bridge between vision and tactile during training. To enhance evaluation in tactile image generation, we propose several reference-free metrics designed to assess the quality, consistency, and diversity of generated tactile images without requiring ground truth comparisons.

\subsection{Tactile Image Generation with Latent Diffusion}

In our framework, we employ a diffusion-based approach \cite{rombach2022high} that progressively refines random noise into coherent samples by reversing a predefined Markov chain. 
A triplet of visual, tactile, and background tactile images \(\{x_{vis}, x_{tac}, x_{bg}\} \in \mathbb{R}^{H \times W \times 3}\) is provided. The background tactile image \(x_{bg}\) records the sensor baseline state in the absence of material contact. To guide training, we incorporate concise textual descriptions \(x_{tex}\) that emphasize essential material-specific patterns—e.g., “rough, bumpy surface typical of concrete.” These cues help the model focus on tactile-relevant features while filtering out irrelevant visual details such as color, lighting or unrelated objects in the scene. While textual cues serve as an essential intermediary between vision and touch during training, our primary goal is to generate a tactile image, \(\hat{x}_{tac}\), solely from the visual input, \(x_{vis}\), during inference. This vision-to-touch approach mirrors the natural human ability to infer tactile properties from visual appearance. Leveraging CLIP’s encoders, both visual and textual inputs are projected into a unified feature space, ensuring consistent representations. As a result, our method synthesizes tactile images using only visual data at inference, as illustrated in Fig.~\ref{fig:diff}.

The tactile image \(x_{tac}\) is first encoded into a compressed latent 
\(\displaystyle z_{tac} = \mathcal{E}(x_{tac}) \in \mathbb{R}^{h\times w\times 3}\)
via a pretrained, frozen VQ-GAN encoder \(\mathcal{E}\) \cite{ding2021vq}. By choosing \(h \times w \ll H \times W\), the aim is to preserve essential information while reducing computational costs. Next, a diffusion step \(t \in \{1, \ldots, T\}\) is sampled uniformly, and 
Gaussian noise is added according to a scheduled variance, resulting in 
the noisy latent \(z_{tac}^{t}\). A time-conditional U-Net \(\epsilon_{\theta}\)
\cite{ronneberger2015u} is trained to predict and remove this noise while 
conditioning on a frozen CLIP text encoder \(\mathcal{C}_{tex}\). The training objective is:
\begin{equation}
  \mathcal{L}(\theta) \;=\; \mathbb{E}_{z_{tac},\,x_{tex},\,\epsilon,\,t}
  \Bigl[
    \|\epsilon_t \;-\; \epsilon_\theta\!\bigl(z_{tac}^{t},\,t,\,\mathcal{C}_{tex}(x_{tex})\bigr)\|_{2}^2
  \Bigr],
\end{equation}
where \(\epsilon_t\) denotes the noise injected at time \(t\). 

Optionally, we incorporate background tactile information by compressing the background image \(x_{bg}\) to obtain
\(z_{bg} = \mathcal{E}(x_{bg})\). We then condition the diffusion process by channel-wise concatenating \(z_{bg}\) 
with the noisy tactile latent \(z_{tac}^{t}\) thereby providing explicit control over the background in the generated tactile image \(\hat{x}_{tac}\) during both training and inference. However, our experiments show that this approach yields inferior performance in generating the desired material-specific tactile patterns.

During inference, the model samples random noise 
\(\displaystyle z_{tac}^{T} \sim \mathcal{N}\!\bigl(0,\, 1\bigr)\)
in latent space. 
The trained model \(\epsilon_{\theta}\) then denoises this latent 
iteratively, stepping backward from \(t=T\) down to \(t=0\). Note that $z_{tac}^0$ represents the fully denoised final latent. Since textual cues are used only during training, the model conditions visual images with CLIP image encoder \(\mathcal{C}_{\mathrm{vis}}(x_{vis})\), combined with 
an unconditional pass via classifier-free guidance for better quality and diversity. The final latent $z_{tac}^0$ is then reconstructed by VQ-GAN decoder \(\mathcal{D}\), 
producing generated tactile image $\hat{x}_{tac} = \mathcal{D}(z_{tac}^0)$.

\subsection{Reference-Free Tactile Generation Metrics}
Fréchet Inception Distance (FID) \cite{heusel2017gans} evaluates generative models by comparing feature distributions, but it assumes a single Gaussian for high-dimensional features—overlooking complexities like multimodality or skewness—and requires around 50,000 images \cite{heusel2017gans} for stable estimates, poses a significant challenge for domains like tactile sensing where data is inherently limited. In contrast, CLIP Maximum Mean Discrepancy (CMMD) \cite{jayasumana2024rethinking} addresses these issues by using MMD which directly compares features without restrictive assumptions, offering a sample-efficient, unbiased, and more reliable evaluations. 
However, CMMD relies on CLIP embeddings, which, despite their strong semantic alignment for natural images, are not trained on tactile data. Consequently, the extracted features fail to capture the material-specific tactile patterns essential for accurate evaluation, thereby limiting the effectiveness of CMMD for tactile generation tasks.

To establish a robust evaluation metric for comparing tactile distributions, a dedicated touch encoder is required to effectively capture the underlying structure of tactile data. We achieve this by pre-training a cross-modal visual-tactile encoder using self-supervised contrastive learning. Building on prior work in contrastive pre-training \cite{tian2020contrastive, fu2024touch}, we employ tactile-textual, vision-textual and tactile-vision contrastive losses to align tactile features with other modalities. To facilitate this alignment, the CLIP image encoder is randomly initialized and repurposed as the tactile encoder $\mathcal{C}_{\mathrm{tac}}$, ensuring a modality-aware representation of tactile data.

\subsubsection{\textbf{Tactile Maximum Mean Discrepancy (TMMD)}} In the context of tactile image evaluation, given two sets of feature embeddings $G = \{\mathbf{g}_1, \mathbf{g}_2, \dots, \mathbf{g}_m \}$ and
$R = \{\mathbf{r}_1, \mathbf{r}_2, \dots, \mathbf{r}_n\}$ where $G$ denotes the generated tactile image features and $R$ corresponds to the reference tactile image features (we discuss the reference-free variants shortly). Building on CMMD’s MMD-based approach, TMMD replaces CLIP embeddings with a dedicated tactile encoder $\mathcal{C}_{\mathrm{tac}}$, ensuring more accurate capture of material-specific tactile nuances. Formally, TMMD is defined as:

\begin{equation}
\begin{aligned}
\widehat{\text{dist}}_{\text{TMMD}}^2(G, R) &= 
\frac{1}{m(m-1)} \sum_{i=1}^{m} \sum_{\substack{j=1 \\ j \neq i}}^{m} k(\mathbf{g}_i, \mathbf{g}_j) \\
&\hspace{-5em} + \frac{1}{n(n-1)} \sum_{i=1}^{n} \sum_{\substack{j=1 \\ j \neq i}}^{n} k(\mathbf{r}_i, \mathbf{r}_j) 
- \frac{2}{mn} \sum_{i=1}^{m} \sum_{j=1}^{n} k(\mathbf{g}_i, \mathbf{r}_j).
\end{aligned}
\end{equation}
where $k$ denotes a positive definite kernel measuring similarity between feature embeddings. Following \cite{jayasumana2024rethinking}, the Gaussian RBF kernel 
$k(\mathbf{x}, \mathbf{y}) = \exp\left(-\|\mathbf{x} - \mathbf{y}\|^2 / 2\sigma^2\right)$ 
is used to ensure TMMD effectively captures the distributional discrepancy in a high-dimensional space. Leveraging $\mathcal{C}_{\mathrm{tac}}$, this metric provides a robust comparison of tactile representations while effectively capturing complex textural details and subtle deformation patterns in the feature space. A lower TMMD score indicates better alignment between the generated and reference distributions, signifying higher quality and fidelity in the generated tactile images. Building on this foundation, we introduce reference‐free metrics in the following sections to further evaluate the generated tactile images without relying on reference data.

\begin{figure}[t]
    \centering
    \includegraphics[width=0.85\linewidth]{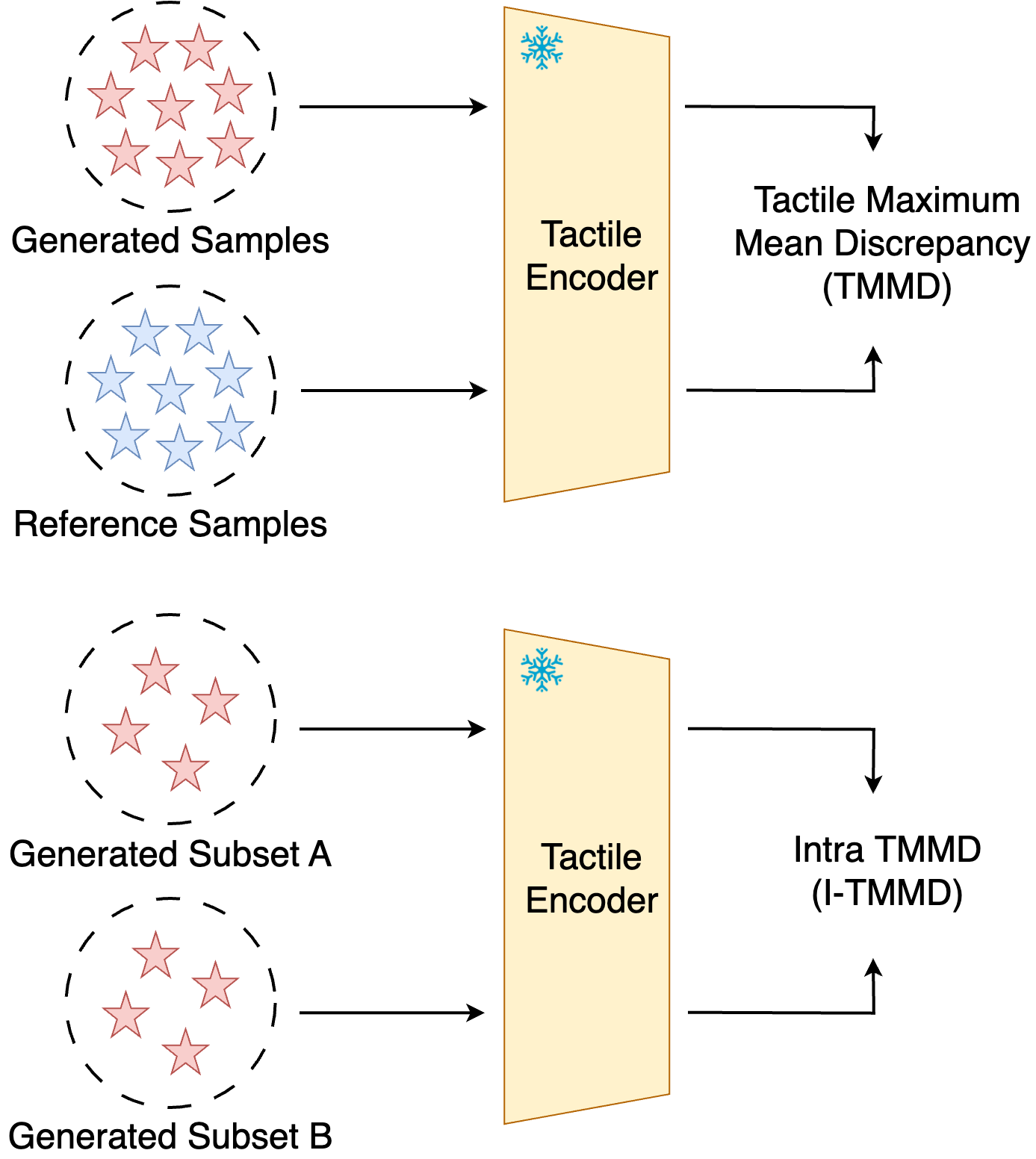}
    \caption{TMMD compares generated and reference tactile features via a dedicated tactile encoder, while I-TMMD provides a reference-free measure of internal consistency by comparing two disjoint subsets of generated samples.}
    \label{fig:metrics}
    
\end{figure}

\subsubsection{\textbf{Intra TMMD (I-TMMD)}}
In generative modeling, obtaining large and representative reference datasets for computing evaluation metrics is often challenging, particularly in specialized domains such as art generation, medical imaging, or tactile generation. These fields frequently suffer from data scarcity and high acquisition costs to obtain high-quality reference distributions. In such scenarios, reference-free evaluation methods become indispensable. To address this need, we introduce I-TMMD as a metric to quantify the internal consistency of generated distributions as illustrated in Fig.~\ref{fig:metrics}. Specifically, I-TMMD operates by splitting the set of generated samples $\mathbf{G}$ into two disjoint subsets $\mathbf{G}_1$ and $\mathbf{G}_2$ and computing the TMMD distance between them: 
\begin{equation}
\text{I-TMMD}(\mathbf{G}) = \widehat{\text{dist}}_{\text{TMMD}}^2(\mathbf{G}_1, \mathbf{G}_2).
\end{equation}

By evaluating the similarity between subsets, I-TMMD provides insight into the model's internal coherence and stability. We are inspired by self-consistency methods in large language models, where multiple stochastic samples are aggregated via majority voting \cite{wang2022self, wan2024dynamic} to ensure convergence of diverse reasoning paths to the same answer. I-TMMD posits that a well-trained generator, converging to a stable and well-defined distribution, should produce statistically similar, internally consistent outputs across its subsets. Thus, a lower I-TMMD score indicates greater internal consistency, reflecting a stable, well-converged generator.

\subsubsection{\textbf{Class-Aware Intra TMMD (CI-TMMD)}}  

While I-TMMD  measures overall internal consistency by splitting the generated samples into two subsets, it does not account for class-specific variations. This can be problematic in scenarios where the generator produces samples that are  consistent globally but fail to capture the distinct characteristics of individual classes. To address this, CI-TMMD extends I-TMMD by enforcing consistency within each class. Let $\mathbf{G} = \{\mathbf{G}^1, \ldots, \mathbf{G}^C\}$ denote generated tactile features partitioned into $C$ classes. For each class $c$, $\mathbf{G}^c$ is split into disjoint equal subsets $\mathbf{G}_1^c$ and $\mathbf{G}_2^c$. CI-TMMD computes the average TMMD distance across classes:
\begin{equation}
\text{CI-TMMD}(\mathbf{G}) = \frac{1}{C}\sum_{c=1}^{C} \widehat{\text{dist}}_{\text{TMMD}}^2(\mathbf{G}_1^c, \mathbf{G}_2^c),
\end{equation}
ensuring the generator preserves class-specific characteristics rather than merely achieving global uniformity.

Despite their utility, I-TMMD and CI-TMMD share a key limitation: they do not guarantee diversity. In cases of mode collapse—where the generator produces similar outputs across classes—the resulting subsets ($\mathbf{G}_1$ and $\mathbf{G}_2$ or $\mathbf{G}_1^c$ and $\mathbf{G}_2^c$) will yield artificially low scores despite catastrophic collapse. Thus, while these metrics effectively assess internal consistency, they should be paired with a metric that evaluates diversity.
 
\subsubsection{\textbf{Diversity-Aware TMMD (D-TMMD)}}  

To ensure classes remain separable, we propose D-TMMD, a reference-free metric that captures both intra-class similarity and inter-class divergence, complementing the internal consistency evaluations of I-TMMD and CI-TMMD. 

For a generator producing outputs partitioned into \(C\) classes \(\mathbf{G} = \{\mathbf{G}^1, \ldots, \mathbf{G}^C\}\), we construct a divergence matrix \(\mathbf{D} \in \mathbb{R}^{C \times C}\) as follows:  

1. \textbf{Intra-Class Consistency}: For diagonal entries (\(c = c'\)), split \(\mathbf{G}^c\) into two subsets \(\mathbf{G}_1^c\) and \(\mathbf{G}_2^c\), then compute:  

\begin{equation}
\mathbf{D}_{c,c} = \widehat{\text{dist}}_{\text{TMMD}}^2(\mathbf{G}_1^c, \mathbf{G}_2^c).
\end{equation}

2. \textbf{Inter-Class Divergence}: For off-diagonal entries (\(c \neq c'\)), compute the divergence between full class sets:  

\begin{equation}
\mathbf{D}_{c,c'} = \widehat{\text{dist}}_{\text{TMMD}}^2(\mathbf{G}^c, \mathbf{G}^{c'}).
\end{equation} 

The D-TMMD metric aggregates these values into a normalized measure of diversity:  

\begin{equation}
\text{D-TMMD}(\mathbf{G}) = \frac{1}{C} \sum_{c=1}^{C} \frac{\mathbf{D}_{c,c}}{\sum_{c'=1}^C \mathbf{D}_{c,c'}}.
\end{equation}  

Here, \(\mathbf{D}_{c,c}\) measures intra-class consistency, while \(\sum_{c'=1}^C \mathbf{D}_{c,c'}\) captures the total divergence of class \(c\) from all classes. A low D-TMMD score indicates strong diversity, as intra-class consistency is small relative to inter-class divergence. Conversely, a high score suggests poor diversity, signaling potential mode collapse or class confusion.  

\begin{figure*}[t]
    \centering
    \includegraphics[width=1\linewidth]{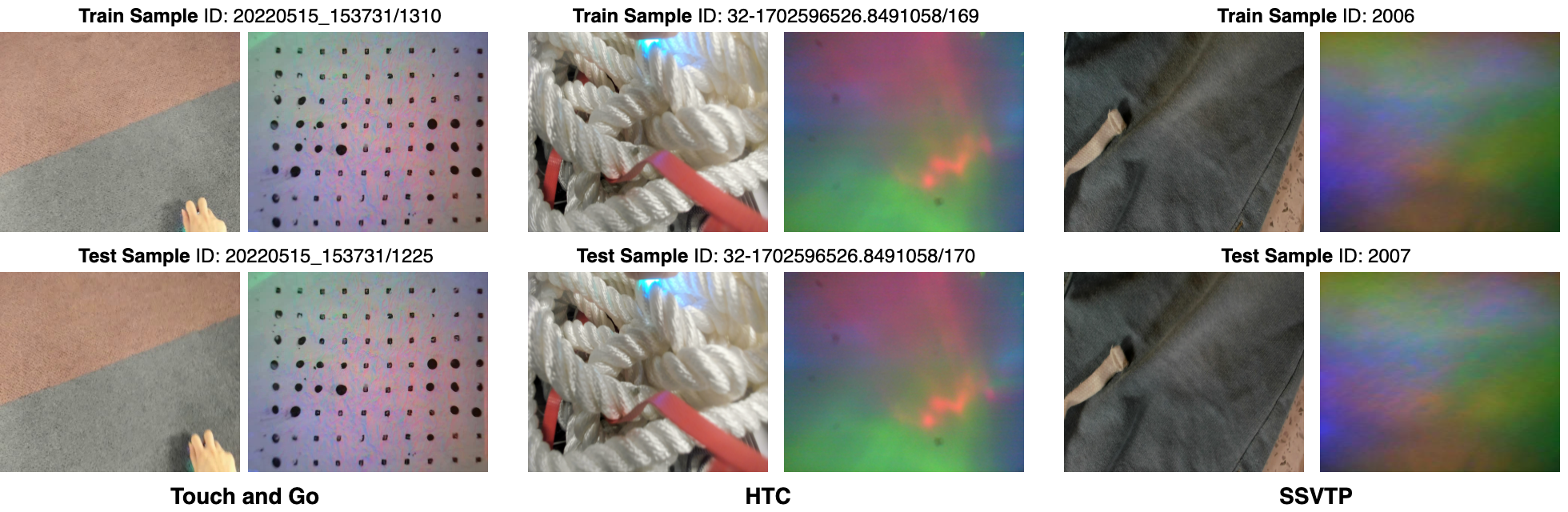}
    
    \caption{Samples from both training and test sets of the Touch and Go, HCT and SSVTP datasets. The near-identical appearance of the vision-tactile pairs highlights the severe data leakage present in the original splits.}
 
    \label{fig:leak}
\end{figure*}

\begin{table*}[t]
\centering
\parbox{.46\linewidth}{
\resizebox{\linewidth}{!}{
\begin{tabular}{lc|cccc}
\toprule
 & \multicolumn{1}{c|}{Material} & \multicolumn{4}{c}{Tactile Generation } \\
 Dataset & Acc ($\%$) & TMMD ($\downarrow$) & I-TMMD ($\downarrow$) & CI-TMMD ($\downarrow$) & D-TMMD ($\downarrow$)\\
\midrule 
\textsc{TaG-Orig} &  54.7 & 0.33 & 0.04 & 0.21 & 0.06\\
\textsc{TaG-NoLeak} & 37.8 & 1.37 & 0.15 & 1.43 & 0.44 \\
\bottomrule
\end{tabular}
}
\caption{Comparison of material classification and tactile generation performance across \textsc{TaG-Orig} and \textsc{TaG-NoLeak}. The results highlight a significant inflation of metrics due to data leakage in \textsc{TaG-Orig}.} 
\label{Table:tag-leak}
}
\hfill
\parbox{.50\linewidth}{
\resizebox{\linewidth}{!}{
\begin{tabular}{lcc|cc|cc}
\toprule
 & \multicolumn{2}{c|}{Tactile-Text Retrieval} & \multicolumn{2}{c|}{Tactile-Vision Retrieval } & \multicolumn{2}{c}{Tactile Generation} \\
 Dataset & Top-1 ($\%$) & Top-5 ($\%$) & Top-1 ($\%$) & Top-5 ($\%)$ & TMMD ($\downarrow$) & I-TMMD ($\downarrow$)\\
\midrule 
\textsc{HCT-Orig} & 36.7 & 70.3 & 79.5 & 95.7 & 0.46 & 0.13 \\
\textsc{HCT-NoLeak} & 4.1 & 17.4 & 7.9 & 27.4 & 2.12 & 0.39\\
\bottomrule
\end{tabular}
}
\caption{Comparison of tactile-text and tactile-vision retrieval accuracy along with tactile generation performance on the HCT dataset. The results indicate a noticeable performance drop on the \textsc{HCT-NoLeak}, underscoring the impact of eliminating data leakage.}
\label{Table:htc-leak}
}

\end{table*}

\section{Results}
\label{sec:results}

\textbf{Implementation details.} In our framework, we first pre-train a dedicated touch encoder following \cite{fu2024touch}. Specifically, we leverage a frozen, pretrained CLIP vision encoder (utilizing a ViT backbone \cite{dosovitskiy2020image}) alongside the CLIP text encoder, while initializing the touch encoder randomly. These encoders are jointly optimized using contrastive learning on normalized latent embeddings with the InfoNCE loss \cite{oord2018representation}. The contrastive training is conducted for 200 epochs with a batch size of 256 on an A100 GPU. This pre-training process is conducted separately for each dataset.

For the latent diffusion model, we follow Stable Diffusion \cite{wang2018high} using the Adam optimizer with a base learning rate of \(2 \times 10^{-6}\). The model is trained for 30 epochs under this learning rate with a batch size of 32 with 2 A100 GPUs, again performed separately on each dataset.  
The CLIP vision and CLIP text are used as conditional encoders and remain frozen throughout diffusion model training. Latent representations are obtained using a frozen, pretrained VQGAN \cite{ding2021vq}, yielding feature maps with spatial dimensions \(64 \times 64\). During inference, the model performs a denoising process over 200 steps with a guidance scale \(s=7.5\).

\noindent
\textbf{Datasets.} 
We use two diverse visuo-tactile datasets:

\begin{itemize}

    \item \textbf{Touch and Go} \cite{yang2022touch} is a comprehensive real-world visuo-tactile dataset capturing human interactions with objects in both indoor and outdoor environments using a GelSight sensor. It comprises 13,900 image-touch pairs, representing approximately 3,971 unique objects across 20 material categories, including fabric, concrete, grass, and wood. In our experiments, we refer to the original splits provided by \cite{yang2022touch} as \textbf{TaG-Orig}. However, we identified that these splits include samples from the same video in both the training and test sets, leading to data leakage. To address this, we created new splits, named \textbf{TaG-NoLeak}, where all data from a single video is exclusively allocated to either the training or test set, ensuring no overlap. Since the dataset lacks textual descriptions, we leverage Molmo \cite{deitke2024molmo}, a large vision-language model, to generate text descriptions for each sample that capture tactile-relevant visual properties.

    \item \textbf{Human Collected Tactile (HCT)} \cite{fu2024touch} is an in-the-wild visuo-tactile dataset collected using a handheld, 3D-printed device with DIGIT sensor that synchronously records visual and tactile data. It comprises 39,154 pairs of in-contact frames with tactile texture descriptions. In our experiments, we denote the original splits provided by \cite{fu2024touch} as \textbf{HCT-Orig}. However, similar to the Touch and Go dataset, these splits include overlapping samples from the same video sequences. We introduced an alternative split, \textbf{HCT-NoLeak}, which assigns all frames from each video exclusively to either the training set or the test set.

\end{itemize}

\noindent
\textbf{Baselines.} We compare our approach to GVSR \cite{yang2023generating}, which uses only visual inputs for training and inference, to showcase the novelty of our method that incorporate textual information during training. 
We also introduce \textsc{Ours w/ BG} which inputs background images \(x_{bg}\) (referred to as ``gel status" in \cite{tu2024texttoucher}) following to prior work \cite{tu2024texttoucher,dou2024tactile}.

\subsection{Data Leakage in Visual-Tactile Datasets} 

Our analysis of the Touch and Go \cite{li2019connecting}, HCT \cite{fu2024touch}, and SSVTP \cite{kerr2022self} datasets reveals systematic data leakage due to temporal proximity between training and test samples, as shown in Fig.~\ref{fig:leak}. These datasets, collected through continuous video recordings of object interactions, contain highly similar consecutive frames. HCT exhibits the most severe leakage, with test frames directly following training samples (e.g., frame 170 in testing after frame 169 in training in Fig.~\ref{fig:leak}), creating near-duplicate pairs. Touch and Go avoids consecutive frames but collects multiple samples from the same region over 3-5 seconds ($\approx$150 frames at 30 fps), leaving test samples as few as 85 frames apart from training data (e.g., frame 1225 vs. frame 1310 in Fig.~\ref{fig:leak}), which is insufficient for meaningful distinction. This overlap persists across numerous samples, leading to leakage. Although SSVTP \cite{kerr2022self} uses a robotic arm with discrete sampling, it still includes highly similar samples across splits. Moreover, a more recent TaRF dataset \cite{dou2024tactile} follows a similar data collection paradigm as Touch and Go \cite{li2019connecting}, suggesting potential leakage risks, though verification remains pending public release of their splits.

\begin{table}[t]
\centering{
\resizebox{\linewidth}{!}{
\begin{tabular}{l|cc|cc}
\toprule
& \multicolumn{2}{c|}{Comparison 1} & \multicolumn{2}{c}{Comparison 2} \\
Metrics & \textsc{Ours} & \textsc{GVST} \cite{yang2023generating} & \textsc{Ours} & \textsc{Ours w/ BG} \\

\midrule 
SSIM ($\uparrow$) &\bf{0.49}& \bf{0.49} & 0.49 & \bf{0.54}\\
PSNR ($\uparrow$) & \bf{15.5} & 15.4 & 15.5 & \bf{16.0}\\
LPIPS ($\downarrow$) & 0.53 & \bf{0.51} & 0.53 & \bf{0.43}\\

\midrule 
FID ($\downarrow$) & 43.9 & \bf{43.6} & \bf{43.9} & 46.9\\
CMMD ($\downarrow$) &2.52& \bf{2.49} & 2.52 & \bf{2.44}\\
\midrule 
TMMD ($\downarrow$) & \bf{1.37} & 1.98 & \bf{1.37} & 2.39\\
I-TMMD ($\downarrow$) & \bf{0.15} & 0.77 & \bf{0.15} & 1.24\\
CI-TMMD ($\downarrow$) & \bf{1.43} & 2.97 & \bf{1.43} & 5.33\\
D-TMMD ($\downarrow$) & \bf{0.44} & 0.92 & \bf{0.44} & 1.40 \\
\midrule 
Human Pref ($\uparrow$) & \bf{75.0\%}  & 25.0\% & \bf{82.5\%} & 17.5\% \\
\bottomrule
\end{tabular}
}
}
\caption{Comparison of methods (\textsc{Ours} vs.\ GVST \cite{yang2023generating} and \textsc{Ours} vs. \textsc{Ours w/ BG}) using both metrics and human preference in \textsc{TaG-NoLeak} dataset. While the newly introduced TMMD, I-TMMD, CI-TMMD and D-TMMD metrics consistently favor the same method that humans prefer, traditional pairwise (SSIM, PSNR, LPIPS) and distribution-based (FID, CMMD) metrics often diverge from human preference. Within each comparison block, the best performer for each metric is in  \bf{bold}.}
\label{Table:human}

\end{table}

The analysis in Tables ~\ref{Table:tag-leak} and ~\ref{Table:htc-leak} underscores the impact of data leakage on performance metrics for different tasks in visual-tactile datasets. Table ~\ref{Table:tag-leak} shows material classification accuracy declining from $54.7$ in \textsc{TaG-Orig} to $37.8$ in \textsc{TaG-NoLeak} ($30.9\%$ drop), and tactile generation errors escalating, with TMMD scores increasing (thus performance decreasing) from $0.33$ to $1.37$, I-TMMD from $0.04$ to $0.15$, CI-TMMD from $0.21$ to $1.43$ and D-TMMD from $0.06$ to $0.44$. Similarly in Table ~\ref{Table:htc-leak}, the original split \textsc{HCT-Orig} exhibits highly inflated retrieval accuracies due to leakage, with tactile-text top-1 retrieval accuracy plummeting from $36.7$ to $4.1$ ($88.8\%$ drop) and tactile-vision top-1 retrieval accuracy dropping from $79.5$ to $7.9$ ($90.1\%$ drop) in the corrected \textsc{HCT-NoLeak} split. Tactile generation errors rise, with TMMD from $0.46$ to $2.12$ and with I-TMMD metrics from $0.13$ to $0.39$. These dramatic performance drops reveal that leakage inflates metrics in the original splits, underscoring the urgent need for rigorous data splitting—which we address by proposing leakage-free datasets—for trustworthy tactile model evaluation.

\begin{figure}[t]
    \centering
    \vspace{-0.1cm} \includegraphics[width=0.7\linewidth]{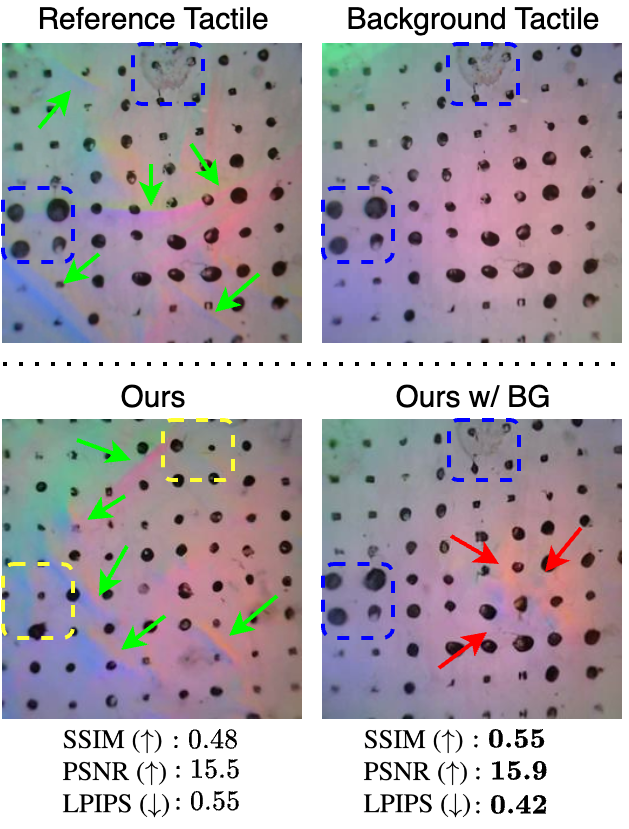}
    \vspace{-0.2cm}
    \caption{This figure compares \textsc{Ours} and \textsc{Ours w/ BG} using traditional pairwise metrics (SSIM, PSNR, LPIPS) to evaluate similarity to the Reference Tactile. \textsc{Ours} accurately captures material-specific features, such as line-shaped patterns for ``grass" (green arrows), while \textsc{Ours w/ BG} incorporates irrelevant background details (blue boxes) using Background Tactile and introduces errors, like pebbly patterns resembling ``concrete" (red arrows). Despite inflated higher performance for \textsc{Ours w/ BG}, these metrics reward irrelevant background details, underscoring their limitations, as \textsc{Ours} better prioritizes relevant material-specific details.}
    \label{fig:bg}  
\end{figure}

\begin{figure*}[t]
    \centering
    \includegraphics[width=1\linewidth]{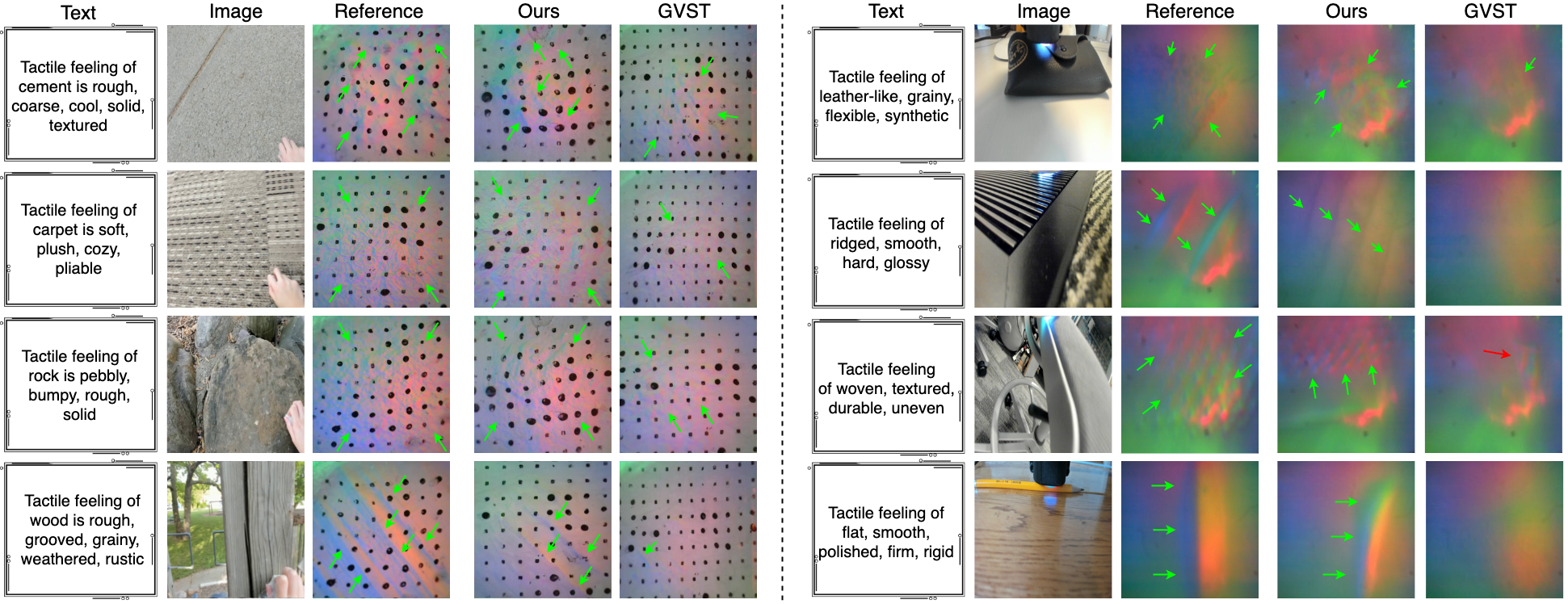}    
    \caption{Qualitative comparison of tactile images generated on the \textsc{TaG-NoLeak} (left) and \textsc{HCT-NoLeak} (right) datasets. Although both methods rely solely on visual inputs during inference, \textsc{Ours}—by incorporating textual descriptions during training—more accurately reproduces the material-specific contact patterns compared to \textsc{GVST} \cite{yang2023generating}. Green arrows highlight desired material-specific patterns.} 
    \label{fig:quali}    
\end{figure*}

\subsection{Novel Metrics and Human Evaluation} 

Next, we validate our newly introduced metrics, in the \textsc{TaG-NoLeak} setting. We conducted two comparisons: (1) \textsc{Ours} vs \textsc{GVST}, and (2) \textsc{Ours} vs \textsc{Ours w/ BG}, in Table ~\ref{Table:human}. We show both traditional and our novel metrics. 
We also report the human preferences for one or the other method's outputs. For this human evaluation, we randomly selected 40 samples from the test set for each method, primarily covering 13 of the most common material categories to ensure a representative selection. Two human raters were asked, in each trial, to choose the generated tactile image that most closely resembled the reference tactile image. They were asked to focus on material-specific patterns—such as line-shaped/elongated textures for grass, pebbly patterns for concrete, and woven/repetitive structures for fabric—while not focusing on sensor’s default background details (highlighted by blue boxes in Fig.~\ref{fig:bg}) that appear even when a sensor does not have any contact with a material. The final preference scores were obtained by averaging the ratings from both raters, who exhibited an $80\%$ agreement in their selections. While $40$ samples sufficed for pairwise metrics (SSIM, PSNR and LPIPS) to align with human evaluation, the statistical demands of FID and MMD-based metrics—requiring larger sample sizes for stable and reliable estimates—necessitated using the entire test set, as smaller sample sizes like $40$ leads to high variance by poorly representing the underlying distributions.

Our findings reveal a strong preference for \textsc{Ours} over both \textsc{GVST} and \textsc{Ours w/ BG}: \textsc{Ours} outperforms \textsc{GVST} by a substantial margin ($75\%$ vs. $25\%$) and surpasses \textsc{Ours w/ BG} even more decisively ($82.5\%$ vs. $17.5\%$). Notably, this preference is clearly captured by our proposed TMMD-based metrics, which consistently give lower (better) scores to \textsc{Ours} (e.g., TMMD drops from $1.98$ with \textsc{GVST} to $1.37$ with \textsc{Ours}, and from $2.39$ with \textsc{Ours w/ BG} to $1.37$ with \textsc{Ours}), aligning closely with human ratings. In contrast, 
FID and CMMD exhibit only marginal differences—sometimes even favoring \textsc{GVST} or \textsc{Ours w/ BG}—thereby failing to reflect the strong human preference for \textsc{Ours}. This discrepancy arises because FID and CMMD rely on RGB-trained encoders, limiting their ability to capture tactile-specific features.

\subsection{Limitations to Pairwise Metrics} 

Although pairwise metrics (focusing on pairs of individual samples, not distributions) such as SSIM, PSNR, and LPIPS are adopted in tactile generation \cite{yang2023generating, tu2024texttoucher, dou2024tactile}, they are inadequate for meaningful evaluation because a substantial portion of a tactile image’s appearance is dictated by sensor-specific background patterns. As shown in Figure~\ref{fig:bg}, \textsc{Ours w/ BG} excels at reproducing these background details (illustrated with blue boxes), misleadingly inflating pairwise metrics despite having little relevance to the contacted material. In contrast, \textsc{Ours} prioritizes the crucial, material-specific contact patterns (green arrows in Fig.~\ref{fig:bg}) but often shows worse performance on pairwise metrics because it does not replicate the sensor’s default background details. Consequently, focusing on background artifacts yields artificially better performance on SSIM, PSNR, and LPIPS even when actual material characteristics are inadequately captured. Indeed, our reference-free metrics (I-TMMD, CI-TMMD, D-TMMD) employ a dedicated tactile encoder to capture material-specific features. 
Further, conditioning the model with background image in \textsc{Ours w/ BG} impedes the learning of relevant tactile features (red arrows highlights inaccurate generation in Fig.~\ref{fig:bg}), as the model relies on replicating  background patterns. As a result, these metrics fail: they often pick the outputs that human evaluators dis-prefer (Table \ref{Table:human}). They also give very similar scores across methods, indicating low discrimination ability.

\begin{table}[t]
{
\centering
\resizebox{\linewidth}{!}{
\begin{tabular}{lcccc|cc}

\toprule
 & \multicolumn{4}{c|}{\textsc{TaG-NoLeak}} & \multicolumn{2}{c}{\textsc{HCT-NoLeak} } \\
 Method & TMMD ($\downarrow$) & I-TMMD ($\downarrow$) & CI-TMMD ($\downarrow$) & D-TMMD ($\downarrow$) & TMMD ($\downarrow$) & I-TMMD ($\downarrow$)\\
\midrule 
\textsc{GVST} & 1.98 & 0.77 & 2.97 & 0.92 & 2.67 & 0.61 \\
\textsc{Ours} & $\mathbf{1.37}$ & $\mathbf{0.15}$ & $\mathbf{1.43}$ & $\mathbf{0.44}$ & $\mathbf{2.12}$ & $\mathbf{0.39}$ \\

\bottomrule
\end{tabular}
}
}
\caption{Quantitative comparison of \textsc{GVST} \cite{yang2023generating} and \textsc{Ours} on \textsc{TaG-NoLeak} and \textsc{HCT-NoLeak} using our proposed metrics.
}
\label{Table:state-art}
\end{table}


\subsection{Comparison with Prior Art}

In Table~\ref{Table:state-art}, we compare \textsc{Ours} with \textsc{GVST} \cite{yang2023generating} on \textsc{TaG-NoLeak} and \textsc{HCT-NoLeak} datasets. On \textsc{TaG-NoLeak} (where class labels are available), \textsc{Ours} lowers TMMD from $1.98$ to $1.37$, I-TMMD from $0.77$ to $0.15$, CI-TMMD from $2.97$ to $1.43$, and D-TMMD from $0.92$ to $0.44$—indicating marked improvements in fidelity, internal consistency, and class separation. Similarly, on \textsc{HCT-NoLeak}, \textsc{Ours} reduces TMMD from $2.67$ to $2.12$ and I-TMMD from $0.61$ to $0.39$, further underscoring the benefit of incorporating textual cues. Overall, our method consistently achieves lower scores, demonstrating its superior ability to preserve material-specific details and generate diverse tactile outputs compared to \textsc{GVST} \cite{yang2023generating}.

Figure~\ref{fig:quali} compares tactile images generated by \textsc{Ours} approach and GVST \cite{yang2023generating} under a leakage-free setting. By incorporating textual descriptions during training, \textsc{Ours} directs the network to focus on tactile-specific features of materials. Thus, even though both methods rely solely on visual inputs during inference, our model more accurately predicts material-specific relevant patterns compared to GVST as illustrated with green arrows in Figure~\ref{fig:quali}. Importantly, our leakage-free setting prevents memorization by ensuring that generated tactile images do not replicate the reference images’ background patterns or the exact locations of material-specific details—features that cannot be inferred solely from visual inputs during inference. 
\section{Conclusion}
\label{sec:conclusion}

We showed that data leakage undermines the generalization of tactile generation models and that conventional metrics fall short in evaluating tactile nuances. Our leakage-free evaluation protocol, coupled with novel reference-free metrics and a text-guided vision-to-touch framework, achieves superior performance and enhanced generalization, paving the way for more reliable tactile sensing systems.

{
\small    
\bibliographystyle{ieeenat_fullname}
\bibliography{main}
}

\end{document}